\documentclass{acmsiggraph}                     

\usepackage[scaled=.92]{helvet}
\usepackage{times}
\usepackage{parskip}
\usepackage[labelfont=bf,textfont=it]{caption}

\usepackage{amsmath}
\usepackage{amsfonts}
\usepackage{algorithm}
\usepackage{algorithmic}
\usepackage{url}

\newcommand{\term}[1]{\emph{#1}}
\newcommand{\dimensions}[2]{$#1\times #2$}
\newcommand{\enumx}[2]{x_{#1},\ldots,x_{#2}}
\newcommand{\RR}{\mathbb{R}}

\DeclareMathOperator{\sign}{sgn}

\title{Fast Weak Learner Based on Genetic Algorithm}
\author{Boris Yangel\\Department of Computational Mathematics and Cybernetics\\Moscow State University, Moscow, Russia\\hr0nix@acm.org}


\begin{document}

\maketitle

\begin{abstract}

An approach to the acceleration of parametric weak classifier boosting is proposed. Weak classifier is called parametric if it has fixed number of parameters and, so, can be represented as a point into multidimensional space. Genetic algorithm is used instead of exhaustive search to learn parameters of such classifier. Proposed approach also takes cases when effective algorithm for learning some of the classifier parameters exists into account. Experiments confirm that such an approach can dramatically decrease classifier training time while keeping both training and test errors small.

\end{abstract}

\keywordlist

\section{Introduction}\label{sec:introduction}

Boosting is one of the commonly used classifier learning approaches. It is machine learning meta-algorithm that iteratively learns additive model consisting of weighed \term{weak} classifiers that belong to some classifier family $W$. In case of two-class classification problem (which we will consider in this paper) boosted classifier usually has form
\begin{equation}
    s(y) = \sign\left(\sum_{i=1}^{N} \alpha_i w_i(y)\right).
\end{equation}
There $y \in Y$ is a sample to classify, $w_i \in W$ are weak classifiers learned during boosting procedure, $\alpha_i$ are weak classifier weights, $w_i(y) \in \{-1, 1\}$, $s(y) \in \{-1, 1\}$. Set $W$ is referred to as \term{weak} classifier family. That is because it elements should have error rate only slightly better than random guessing. It expresses the key idea of boosting: strong classifier can be built on top of many weak.

There are many boosting procedures that differ by the type of loss being optimized for the final classifier. But no matter what kind of boosting procedure is used, on each iteration it should select (learn) a weak classifier with minimal weighed loss from $W$ family using special algorithm called \term{weak learner}. Fast and accurate optimization methods are often not applicable there (especially in the case of discrete classifier parameters), so exhaustive search over weak classifier parameter space is used as a weak learner. Unfortunately, exhaustive search can take a lot of time. For example, learning cascade of boosted classifiers based on haar features with \term{AdaBoost} and exhaustive search over classifier parameter space took several weeks in the famous work \cite{Viola2001Robust}. That's why it is often very important to decrease weak classifier learning time using some appropriate numerical optimization approach.

One of the widely used approaches to the numerical optimization is genetic algorithm \cite{Goldberg1989Genetic}. It is based on biological evolution ideas. Optimization problem solution is coded as \term{chromosome} vector. \term{Initial population} of solutions is created using random number generator. \term{Fitness function} is then used to assign fitness value to every population member. Solutions with the biggest fitness values are selected for the next step. In the next step, \term{genetic operators} (crossover and mutation usually) are applied to selected chromosomes to produce new solutions and to modify existing ones slightly. That modified solutions form up a new generation. Then described process repeats. That's how evolution is modeled. It continues until global or suboptimal solution is found or time allowed for evolution is over. Genetic algorithms are often used for global extremum search in big and complicated search spaces. It makes genetic algorithm good candidate for weak classifier learner.

\section{Related work}\label{sec:related_work}

Usage of genetic algorithm for weak learner acceleration was already proposed in several works. For example, in \cite{Treptow2004Combining} genetic weak learner with special crossover and mutation operators was used to learn classifier based on extended haar feature set. In \cite{Ramirez2007Face} genetic algorithm was used to select a few thousand weak classifiers with smallest error on unweighed training set before boosting process starts. Then exhaustive search over selected classifiers was performed on each boosting iteration to select the one with minimal weighed loss. In \cite{Masada2008GA} boosting procedure was completly integrated with genetic algorithm. Few classifiers were selected on each boosting iteration from solution population and added to the strong classifier. That selected classifiers were then used to produce new population members by applying genetic operators.  Then, in \cite{Abramson2006Combining} authors used for weak learner some special evolutionary algorithm they've called \term{Evolutionary Hill-Climbing}. Crossover operator was not used in it. Instead, $5$ different mutations were applied to every population member on each algorithm iteration. Result of each mutation was rejected when it did not improve fitness function value.

There were two main reasons for using genetic search instead of any other approaches in these works. Most of the classifiers used in mentioned works were some extensions of the haar classifier family originally proposed in \cite{Viola2001Robust}. So, huge size of a weak classifier family do not allow to apply exhaustive search based optimization. And complicated discrete structure of a weak classifier blocks all other optimization options.

Another important observation is the fact that every time work authors were forced to implement some specialized solution for genetic weak learner. So, ability to generalize evolutionary approach to learning weak classifier is investigated in this work.

\section{Proposed method}\label{sec:proposed_method}

We are interested in developing some general approach to learning weak classifier. This approach should work much more faster than exhaustive search over classifier parameter space. In the following document sections one such approach is presented. It is based on the fact that when number of classifier parameters to optimize is fixed, weighed loss optimization problem simply turns out into multivariate function minimization problem which is well-developed area of genetic algorithm application.

\subsection{Population member} \label{subsec:population_member}

Let $W$ be some parametric family of weak classifiers. It means that every weak $w \in W$ can be described by set of it's $n$ real-valued parameters $\enumx{1}{n}$. Let's also assume that for last $l$ parameters ($l$ can be equal to zero) there exists some effective learning algorithm $L_E:\RR^{n-l} \to \RR^l$. We will refer to such parameters as to \term{linked}. For given values of parameters $\enumx{1}{n-l}$, called \term{free},  $L_E$ finds optimal values for linked parameters that minimize loss function $E:\RR^n \to \RR^+$. It means that our task is to find values of free parameters that deliver the minimum to the loss function $E[\enumx{1}{n-l},L_E(\enumx{1}{n-l})]$. So, set of parameters $\enumx{1}{n-l}$ represents solution to our optimization problem and form up a member of genetic algorithm population.

\subsection{Fitness function} \label{subsec:fitness_function}

It is natural to assume that classifier with small error on training set should have greater probability to get to the next generation of genetic algorithm. That allows us to introduce fitness function $F:\RR^{n-l} \to \RR^+$ as follows:
\begin{multline} \label{equation:fitness_function}
    F(\enumx{1}{n-l})=\\
    =1/E[\enumx{1}{n-l},L_E(\enumx{1}{n-l})].
\end{multline}

We do not consider $E=0$ case. Classifier can not be called weak if it has zero error value on training set. If such a classifier is presented in a weak classifier family, we can select only that classifier as a whole boosting procedure result.

\subsection{Genetic representation} \label{subsec:genetic_representation}

 Every approach that allows us to code a set of free parameters is appropriate for population member representation. In this work we have selected binary string representation which was confirmed to be effective in function optimization problems. Some alternative representations can be found, for example, in \cite{Goldberg1989Genetic}.

 To form the binary string classifier representation, each classifier parameter should be first represented as a binary string of fixed length, using fixed-precision encoding. Then all the parameters can be simply concatenated to form the final binary string of fixed length.

Sometimes point $p \in \RR^n$ can have no corresponding classifier. For the different families of image region classifiers it is possible, for example, when one of the free parameters representing top-left corner of a classifier window is below zero. In this case fitness function value for the population member representing that point can be forced to be zero. That is how such situations were dealt with in experiments described in section \ref{sec:experiments}. Another possible approach is to select representation and genetic operators in a way that simply does not allow such points to appear. But that approach is less general.

\subsection{Genetic operators} \label{subsec:genetic_operators}

In this work we've used two most common genetic operators: mutation and crossover. For binary string representation mutation and crossover are usually defined as follows:
\begin{itemize}
    \item Crossover operator selects random position in the binary string. Then it swaps all the bits to the right of the selected position between two chromosomes. Such crossover implementation is called 1-point crossover.
    \item Mutation operator changes value of the random chromosome bit to the opposite.
\end{itemize}

In our case, crossover operator produces two new solutions from the two given chromosomes as following: some of the parameters (placed to the left of the selected position) are taken from the first classifier, some of the parameters (placed to the right) --- from the second. And one parameter, probably, can be made from both the the first and the second classifier. Mutation operator simply produces new solution by changing value of the random classifier parameter.

\subsection{Algorithm summary} \label{subsec:algorithm_summary}

\begin{algorithm}
\caption{Genetic weak learner}\label{algorithm}
\begin{algorithmic}[1]
\STATE Generate initial population of $N$ random binary strings;
\FOR{$i=1$, \ldots, $K_{max}$}
    \STATE Add $\lceil NR_c \rceil$ members to the population by applying crossover operator to the pairs of the best population members;
    \STATE Apply mutation operator to $\lceil NR_m \rceil$ random population members;
    \STATE Calculate value of \eqref{equation:fitness_function} for each population member;
    \STATE Remove all the population members except of the $N$ best (the ones with highest value of \eqref{equation:fitness_function});
\ENDFOR
\RETURN weak classifier associated with point represented by best population member as a result;
\end{algorithmic}
\end{algorithm}

Algorithm \ref{algorithm} uses elitism as a population member selection approach. It has 4 parameters:
\begin{itemize}
    \item $N > 0$ --- population size.
    \item $K_{max} > 0$ --- number of generations.
    \item $R_c \in (0, 1]$ --- crossover rate.
    \item $R_m \in (0, 1]$ --- mutation rate.
\end{itemize}

\subsection{Discussion} \label{subsec:discussion}

Advantage of the proposed method lies in the fact that computational complexity of the weak learner does not depend on the size of the weak classifier family. One can achieve balance between training time and classifier performance only by changing values of $N$, $K_{max}$ and $S$ (discussed later). Similar effect can be achieved by shrinking weak classifier family itself. But in most cases prior knowledge about weak classifier performance in boosting is simply not available.

One of the main disadvantages of the proposed weak learner is the fact that many potentially interesting weak classifiers can not be represented as a parameter vector of constant length. For example, decision trees, widely used in boosting, can have variable number of nodes. Misclassification loss we want to optimize should also be more or less stable as a function of classifier free parameters. If small perturbations of the free parameter vector lead to the unpredictable changes in the loss function value, genetic optimization does not make much sense, becoming just a random search. But, unfortunately, that situation happens quite often, especially if classifier parameter count is small. Common example is a situation when one of the free parameters represents feature number and features with close numbers are not correlated at all.

\section{Experiments} \label{sec:experiments}

\subsection{Algorithms for experiments}

Two boosting-based algorithms were implemented to compare proposed genetic weak learner with original learners proposed by algorithm authors. \term{Viola-Jones} \cite{Viola2001Robust} and \term{Face alignment via boosted ranking model} \cite{Wu2008Face} were selected for that purpose because both algorithms use parametric weak classifiers applied to image regions. These algorithms are based on distinct boosting procedures (\term{AdaBoost} and \term{GentleBoost}), so loss, sample weight and classifier weight functions used in them differ a lot. Another difference between selected algorithms is a problem they solve: two-class classification in \cite{Viola2001Robust} and ranking in \cite{Wu2008Face}. Training time of the naive implementation is quite long for both algorithms, so acceleration of boosting process is necessary.

Weak classifiers used in both algorithms are based on haar features and have common set of adjustable parameters. So, weak classifier in both problems can be represented as $w_i=(x_i, y_i, width_i, height_i, type_i, g_i, t_i)$. There $x_i$, $y_i$, $width_i$ and $height_i$ describe image region, $type_i$ encodes haar feature type, $g_i$ is a haar feature sign and $t_i$ represents weak classifier threshold. Parameters $g_i$ and $t_i$ are linked because both algorithms have an effective algorithm for learning them. Parameter $type_i$ was also made linked: changing feature type during genetic optimization does not make much sense because it can change fitness function value significantly after just one mutation or crossover. Separate algorithm run was performed instead for each feature type. Best result from all the runs was then selected. We've used the same $5$ haar feature types as in \cite{Wu2008Face} for training both classifiers.

\subsection{Run patterns}

Comparison of two different genetic algorithm run patterns was also performed in this work. One pattern considered was running genetic optimization once with big population size. Another pattern used was running optimization algorithm multiple times (denoted as $S$) with small population size and then selecting best found classifier. When population size is small, final solution depends on initial population a lot. So, considerably different results can be obtained for different algorithm runs. While this run pattern produces worse classifiers, it can be implemented on multiprocessor and multicore architectures very efficiently: each processing unit can run it's own genetic simulation. That makes perfect parallel algorithm acceleration possible.

\subsection{Training and test sets}

As in work \cite{Treptow2004Combining}, \cite{CarbonettoViolaJonesData} human faces database was used to train and test classifier for Viola-Jones algorithm. Database was divided in half to form the training and test sets. Each sample has size of \dimensions{24}{24} pixels.

Face images with landmarks from FG-NET aging database were used to form the database for learning face alignment ranker proposed in \cite{Wu2008Face}. 600 face images were selected from database and then resized to size of \dimensions{40}{40} pixels. 400 images were used to produce training set and other 200 --- for testing. 10 sequential 6-step random landmark position perturbations were then applied to selected face images to produce images of misaligned faces, as described in original paper. Training and test set samples were then made of pairs of images with increasing alignment quality.

\subsection{Hardware}

All the experiments were performed on PC equipped with $2.33$ GHz Intel Core $2$ Quad processor and $2$ GB of DDR2 RAM.

\subsection{Results}

\begin{table}[t]
\centering
\caption{Viola-Jones, acceleration}\label{table:violajones:time}
\begin{tabular}{ccccc}
\hline
\multicolumn{3}{c}{Run pattern} & Time (sec)        & Acceleration        \\
\hline
$S$ & $N$   & $K_{max}$         &                   &                     \\
\hline
$1$ & $50$  & $10$              &   $\mathbf{2.82}$ &   $\mathbf{329.38}$ \\
$1$ & $100$ & $20$              &   $9.40$          &   $98.77$           \\
$1$ & $400$ & $40$              &   $100.29$        &   $9.26$            \\
$10$ & $10$ & $20$              &   $4.00$          &   $231.94$          \\
$20$ & $20$ & $40$              &   $28.74$         &   $32.31$           \\
\hline
\multicolumn{3}{c}{Brute force} &   $928.52$        &   $1.00$            \\
\hline
\end{tabular}
\end{table}

\begin{table}[t]
\centering
\caption{Viola-Jones, error}\label{table:violajones:error}
\begin{tabular}{ccccc}
\hline
\multicolumn{3}{c}{Run pattern} & \multicolumn{2}{c}{Error}                         \\
\hline
$S$ & $N$   & $K_{max}$         &   Learning                &   Test                \\
\hline
$1$ & $50$  & $10$              &   $0.0005$                &   $0.0356$            \\
$1$ & $100$ & $20$              &   $0.0002$                &   $0.0380$            \\
$1$ & $400$ & $40$              &   $0.0000$                &   $\mathbf{0.0328}$   \\
$10$ & $10$ & $20$              &   $0.0003$                &   $0.0378$            \\
$20$ & $20$ & $40$              &   $0.0000$                &   $0.0391$            \\
\hline
\multicolumn{3}{c}{Brute force} &   $0.0000$                &   $0.0349$            \\
\hline
\end{tabular}
\end{table}

\begin{table}[t]
\centering
\caption{Face alignment via BRM, acceleration}\label{table:brm:time}
\begin{tabular}{ccccc}
\hline
\multicolumn{3}{c}{Run pattern} & Time (sec)            & Acceleration          \\
\hline
$S$ & $N$   & $K_{max}$         &                       &                       \\
\hline
$1$ & $25$  & $10$              &   $\mathbf{68.15}$    &   $\mathbf{5195.88}$  \\
$1$ & $50$  & $10$              &   $173.33$            &   $2043.09$           \\
$2$ & $75$  & $15$              &   $909.55$            &   $389.34$            \\
$4$ & $100$ & $20$              &   $3582.37$           &   $98.85$             \\
\hline
\end{tabular}
\end{table}

\begin{table}[t]
\centering
\caption{Face alignment via BRM, error}\label{table:brm:error}
\begin{tabular}{ccccc}
\hline
\multicolumn{3}{c}{Run mode}    & \multicolumn{2}{c}{Error}                         \\
\hline
$S$ & $N$   & $K_{max}$         &   Learning                &   Test                \\
\hline
$1$ & $25$  & $10$              &   $0.0278$                & $0.0317$              \\
$1$ & $50$  & $10$              &   $0.0246$                & $0.0297$              \\
$2$ & $75$  & $15$              &   $0.0199$                & $0.0268$              \\
$4$ & $100$ & $20$              &   $\mathbf{0.0173}$       & $\mathbf{0.0259}$     \\
\hline
\end{tabular}
\end{table}

Tables \ref{table:violajones:time} and \ref{table:brm:time} show average duration of $1$ boosting iteration together with comparison to exhaustive search. Tables \ref{table:violajones:error} and \ref{table:brm:error} show error rate of the final classifiers on the training and test sets. We have not trained any classifier using exhaustive search for boosted ranking model because it would take about a year to finish the process on our training set.

Experiments with Viola-Jones object detector showed that classifier trained using genetic weak learner performs only slightly worse than classifier trained using exhaustive search over classifier space. For $N=400$ final classifier even shows better performance. Classifier trained with $S=1$, $N=50$ and $K_{max}=10$ accelerates boosting nearly $300$ compared to exhaustive search times while still performing good on test set. Classifiers trained with small $N$ and big $S$ values (using second run pattern) perform worse than any other. But, as it was mentioned before, such classifiers can be trained on multiprocessor or multicore systems very efficiently.

Experiments with face alignment via boosted ranking model showed how exactly classifier performance depends on values of $S$, $N$ and $K_{max}$. Increasing value of the each parameter results in increased training time, but also in increased classifier performance. Nevertheless, difference in training time is much more significant compared to the difference in prediction error. Classifier with $S=1$, $N=25$ è $K_{max}=10$ was trained 50 times faster than the best obtained classifier for BRM, but it's error is only $1.2$ times worse. It makes such a classifier a perfect candidate for preliminary experiments that usually take place before training final classifier starts.

\section{Conclusion}\label{sec:conclusion}

An approach to boosting procedure acceleration was proposed in this work. Approach is based on usage of special genetic weak learner for learning weak classifier on each boosting iteration. Genetic weak learner uses genetic algorithm with binary chromosomes. That genetic algorithm is designed to solve an optimization problem of selecting weak classifier with the smallest weighed loss from some parametric classifier family. Proposed method was generalized for the case when there exists an effective algorithm for learning some of the parameters of a weak classifier. Experiments have shown that such approach allows us to accelerate training process dramatically for practical tasks while keeping prediction error small.

Genetic weak learner proposed in this work can't be used to boost any tree-based classifiers. That fact limits its usage in many scenarios because stump weak classifiers can not represent any relationships between different object features. So, in the future work we plan to generalize our approach for accelerating tree-based boosting.

Another option for future research is performing additional experiments with classifiers not related to haar features in any way. That will confirm proposed algorithm's profit in computer vision problems not biased towards haar feature usage. In fact, it would be nice to determine different parametric classifier families that can be efficiently boosted using proposed weak learner.

\bibliographystyle{acmsiggraph}
\bibliography{Paper}
\end{document}